\documentclass[journal]{IEEEtran}
\ifCLASSINFOpdf
\else
\fi

\usepackage{cite}
\usepackage{graphicx}
\usepackage{multicol}
\usepackage{subfig}
\usepackage{amsmath}
\usepackage{amsfonts}
\usepackage{romannum}
\usepackage{caption}
\usepackage{tabularx}
\usepackage{tabulary}
\usepackage{array}
\usepackage{tabu}
\usepackage{xcolor}
\usepackage{multirow}

\DeclareMathOperator*{\argmax}{arg\,max}  

\begin{document}

\title{Pose-Aware Instance Segmentation Framework from Cone Beam CT Images for Tooth Segmentation}

\author{Minyoung Chung, Minkyung Lee, Jioh Hong, Sanguk Park, Jusang Lee, Jingyu Lee, \\Jeongjin Lee$^{\ast}$, and Yeong-Gil Shin%
\thanks{\textit{Asterisk indicates corresponding author.}}%
\thanks{M. Chung, M. Lee, J. Hong, S. Park, J. Lee, J. Lee, and Y.-G. Shin are with the Department of Computer Science and Engineering, Seoul National University, Republic of Korea (e-mail: chungmy@snu.ac.kr).}%
\thanks{*J. Lee is with the Department of Computer Science and Engineering, Soong-sil University, Republic of Korea (e-mail: profjjlee@naver.com).}}


\maketitle

\begin{abstract}
   Individual tooth segmentation from cone beam computed tomography (CBCT) images is an essential prerequisite for an anatomical understanding of orthodontic structures in several applications, such as tooth reformation planning and implant guide simulations. However, the presence of severe metal artifacts in CBCT images hinders the accurate segmentation of each individual tooth. In this study, we propose a neural network for pixel-wise labeling to exploit an instance segmentation framework that is robust to metal artifacts. Our method comprises of three steps: 1) image cropping and realignment by pose regressions, 2) metal-robust individual tooth detection, and 3) segmentation. We first extract the alignment information of the patient by pose regression neural networks to attain a volume-of-interest (VOI) region and realign the input image, which reduces the inter-overlapping area between tooth bounding boxes. Then, individual tooth regions are localized within a VOI realigned image using a convolutional detector. We improved the accuracy of the detector by employing non-maximum suppression and multiclass classification metrics in the region proposal network. Finally, we apply a convolutional neural network (CNN) to perform individual tooth segmentation by converting the pixel-wise labeling task to a distance regression task. Metal-intensive image augmentation is also employed for a robust segmentation of metal artifacts. The result shows that our proposed method outperforms other state-of-the-art methods, especially for teeth with metal artifacts. Our method demonstrated 5.68\% and 30.30\% better accuracy in the F1 score and aggregated Jaccard index, respectively, when compared to the best performing state-of-the-art algorithms. The primary significance of the proposed method is two-fold: 1) an introduction of pose-aware VOI realignment followed by a robust tooth detection and 2) a metal-robust CNN framework for accurate tooth segmentation.
\end{abstract}


\IEEEpeerreviewmaketitle

\section{Introduction}
\begin{figure}[t]
    \centering
    \includegraphics[width=\linewidth]{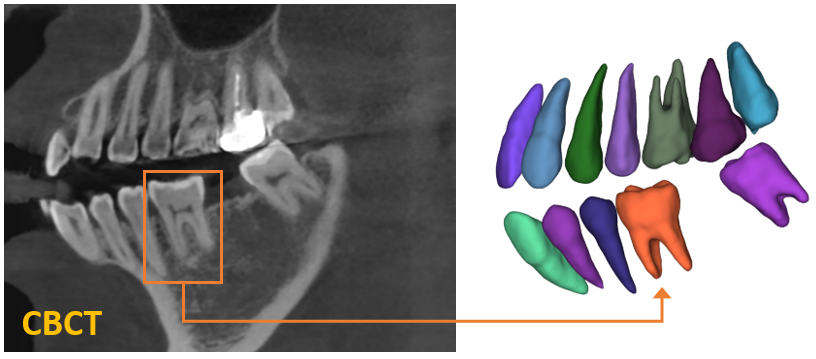}
    \caption{Individual tooth segmentation from a cone beam computed tomography (CBCT) image. Example of surface visualization of tooth segmentation results.}
    \label{fig:intro}
\end{figure}

\IEEEPARstart{D}{igitized} orthodontic applications in dentistry have increased based on the development of cone beam computed tomography (CBCT) imaging. CBCT is a widely used medical imaging technique that provides high resolution 3D volumetric data. To build an effective computer-aided diagnosis system in orthodontic applications, such as oral treatment planning for tooth reformation and implant guide simulation, automatic segmentation of individual teeth from the CBCT images is an essential prerequisite (Fig. \ref{fig:intro}). However, accurate segmentation of an individual tooth from a CBCT image is a challenging task owing to heterogeneous intensity distribution, unclear boundaries between the tooth root and alveolar bone (Fig. \ref{fig:tooth_boundaries_alveolar}), and diverse shapes and poses. Moreover, the majority of CBCT images contain severe metal artifacts that hinder the accurate segmentation of teeth (Fig. \ref{fig:tooth_boundaries_metal}). In this study, we propose a fully automated instance segmentation framework using 3D images of teeth that is robust to several challenging conditions of the teeth, such as, dynamic poses, missing teeth, inter-tooth proximity, and presence of severe metal artifacts.\par

An extensive amount of literature on individual tooth segmentation were proposed in the last few decades. Classical image processing methods that exploited region growing \cite{akhoondali2009rapid}, morphological operations \cite{kakehbaraei2018dental}, and watershed algorithm \cite{fan2019marker, kakehbaraei2018dental} were studied. Several works employed contour-based level-set methods \cite{gao2010individual, ji2014level, gan2017tooth, wang2018accurate} or shape-based registration methods \cite{barone2016ct, pei20163d}. However, all the classical algorithms demonstrated limitations while handling the aforementioned challenging conditions, such as heterogeneous intensities, unclear boundaries, diverse anatomical poses, and presence of metal artifacts. Moreover, classical algorithms typically require manual seed points to perform tooth segmentation, which results in a semiautomated application. More recently, a few studies \cite{miki2017classification, cui2019toothnet} applied convolutional neural network (CNN) architectures to resolve individual tooth segmentation. Although the proposed CNN-based methods showed promising results over previous approaches, the metal artifact condition is still difficult to overcome, implying that accurate detection and segmentation is still a challenging task. Note that the teeth including severe metal artifacts were not previously researched, either in the methodological or in the experimental perspectives.\par

To employ a CNN for individual tooth segmentation, we chose the instance segmentation framework. Unlike other organs in medical imaging, such as liver or spleen, there are multiple instances of a tooth in a single CBCT image. Thus, individual tooth segmentation requires an instance segmentation technique rather than semantic segmentation, which is performed through preceding object detection. The primary challenges of individual tooth detection and segmentation arise from: 1) high overlapping ratio between instances (i.e., teeth) and 2) the existence of severe metal artifacts in CBCT images. These two primary issues degrade the accuracy of the individual tooth segmentation task.\par

In this study, we propose a fully automatic, hierarchical method that performs accurate individual tooth detection followed by a CNN based on single pixel-wise labeling to segment the tooth (Fig. \ref{fig:architecture}). We first extracted the volume-of-interest (VOI) region by pose regression using neural networks. Then, the extracted VOI region was realigned based on the pose (i.e., axes). Subsequently, we detected individual teeth and performed individual tooth segmentation. The key components of our method are the accurate tooth detection framework through VOI realignment, which reduces the inter-overlapping area between boxes, multiclass classification within a detector, which boosts the accuracy of metal-tooth detection, and the metal-robust CNN for accurate tooth segmentation.\par

The remainder of this paper is structured as follows. In Section \Romannum{2}, we review the related works on tooth segmentation methods and CNN architectures that perform object detection and segmentation. Further, we describe our proposed method in Section \Romannum{3}. Section \Romannum{4} demonstrates experimental results and Section \Romannum{5} presents the discussion and conclusion.

\section{Related Works}
In this section, we first review the literature on tooth segmentation methods that were proposed until recently. In the following subsections, the CNN architectures for object detection and medical image segmentation will be highlighted.


\subsection{Literature on Tooth Segmentation}
\subsubsection{Classical Methods}
Classical image processing methods have been widely studied to achieve tooth segmentation \cite{hiew2010tooth, kakehbaraei2018dental, wang2016automated, fan2019marker}. Several methods including region growing \cite{akhoondali2009rapid}, watershed algorithm \cite{kakehbaraei2018dental, fan2019marker}, morphological operators \cite{kakehbaraei2018dental}, graph-cut-based segmentation \cite{hiew2010tooth}, template-based registration \cite{barone2016ct, pei20163d}, and random forest classification \cite{wang2016automated} were implemented. Semiautomatic algorithms with manually annotated cues for easy implementation have also gained popularity \cite{zou2017semi, fan2019marker, zhao2006interactive}. The primary drawbacks of the classical methods are their reliance on the intensity or anatomical heuristics (i.e., assumptions). The algorithms break down in many cases where patient conditions do not satisfy the assumptions, such as intensity dynamics on metal-teeth or large anatomical shape variations (e.g., missing or misalignment of teeth).

\begin{figure}[t]
    \centering
    \subfloat[Boundaries between the tooth root and alveolar bone.]{
    \includegraphics[width=\linewidth]{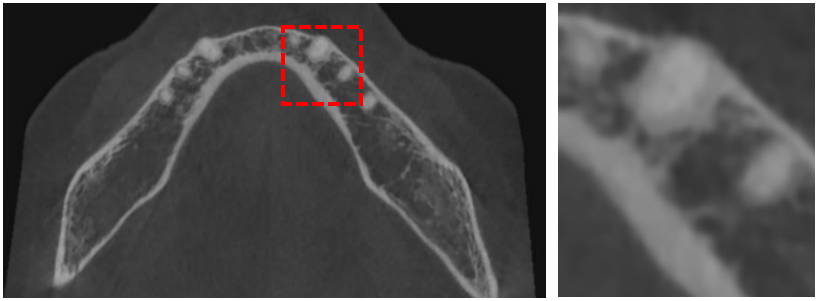}
    \label{fig:tooth_boundaries_alveolar}}
    
    \centering
    \subfloat[Boundaries in metal artifact region.]{
    \includegraphics[width=\linewidth]{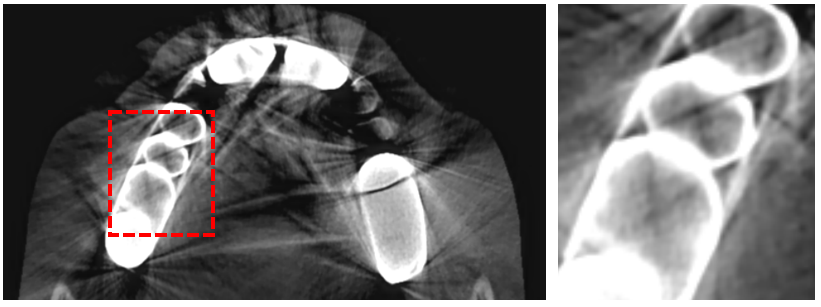}
    \label{fig:tooth_boundaries_metal}}
    
    \caption{Unclear boundaries in CBCT images.}
    \label{fig:tooth_boundaries}
\end{figure}

\subsubsection{Level-set Methods}

The application of a level-set method to the individual tooth segmentation was extensively studied \cite{gao2010individual, ji2014level, gan2017tooth, xia2017individual, wang2018accurate}. In \cite{gao2010individual}, two different level sets were introduced to handle the dynamic root branching and crown parts. The authors applied a coupled level-set method to resolve the adjacency problem between neighboring crowns \cite{gao2010individual}. The intensity distribution was studied to enhance the performance of a level-set framework that models the inside and outside of the tooth \cite{ji2014level}. To tackle the over-segmentation problem presented by the proximity of alveolar bone, simultaneous segmentation of tooth and alveolar bone was proposed \cite{gan2017tooth}. The hybrid level-set model was presented in \cite{wang2018accurate} by forming both local likelihood image fitting and prior shape constraint energy terms. The authors regularized the level-set functional by a reaction diffusion \cite{wang2018accurate}.\par

Despite all the appealing features from the contour propagating scheme, level-set methods demonstrate a common limitation, i.e., it is difficult to delineate unclear boundaries (Fig. \ref{fig:tooth_boundaries}). In a CBCT, the boundaries between the tooth root and the alveolar bone have weak edge characteristics (i.e., a low magnitude of gradient; Fig. \ref{fig:tooth_boundaries_alveolar}) and the presence of severe metal artifacts (Fig. \ref{fig:tooth_boundaries_metal}), which hinder the accurate propagation of a contour-based level-set method. Furthermore, level-set methods require tedious manual user interaction to define the initial contour \cite{gao2010individual, ji2014level, wang2018accurate}, indicating a semiautomated algorithm.\par

\subsubsection{Convolutional Neural Network (CNN)-based Methods}
The body of literature focusing on CNNs for the individual tooth segmentation task is relatively low. Recently, a full-CNN-based method was presented that performs individual tooth segmentation \cite{cui2019toothnet}. The authors employed instance segmentation architecture using an end-to-end CNN framework similar to the Mask region-based CNN (Mask-RCNN) \cite{he2017mask}. The singularities of the proposed method were the employment of edge map, similarity matrix, and combined tooth identification to boost the performance. However, metal artifacts, which are commonly used in dental clinics, were not considered in their proposed method.

\begin{figure*}[t!]
    \centering
    \subfloat[Architecture of the volume-of-interest (VOI) extraction and realignment. The input CBCT is first projected to a 2D image based on the maximum intensities. Point and line pairs are regressed by 2D convolutional neural network (CNN) for pose extraction. Finally, the two VOIs are realigned by fixed margins. We applied y-axis flipping for the lower part to align the teeth. A single point and line pair are visualized for simplicity.]{
    \includegraphics[width=\linewidth]{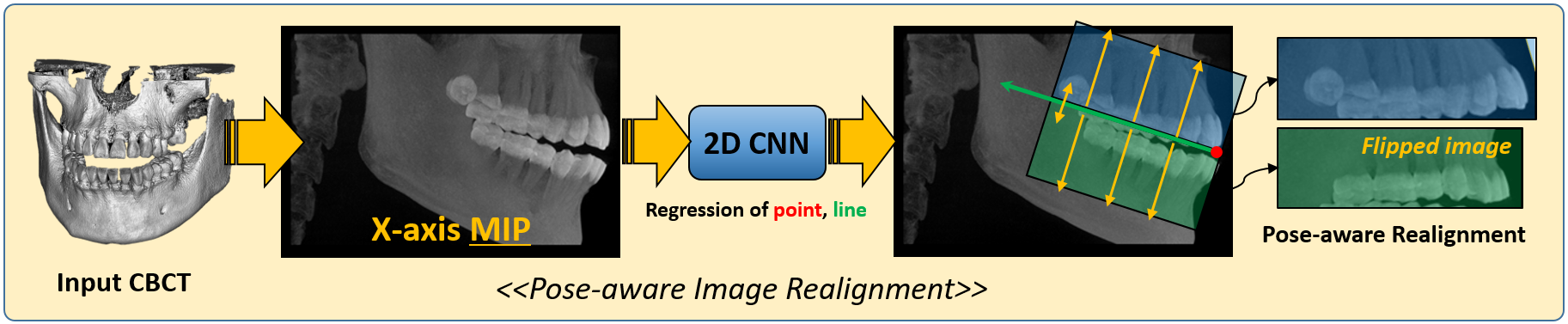}
    \label{fig:archi_align}}

    \centering
    \subfloat[Our proposed instance segmentation framework. We adopted the backbone of the faster region-based CNN \cite{ren2015faster} encoder part and region proposal network. Non-maximum-suppression-based sampling and group-wise multiclass classification are applied to train the detector. The original CBCT image is then cropped according to the detected boxes and fed to the 3D fully convolutional network (FCN) (Fig. \ref{fig:network}) through cutout augmentation to perform individual tooth segmentation.]{
    \includegraphics[width=\linewidth]{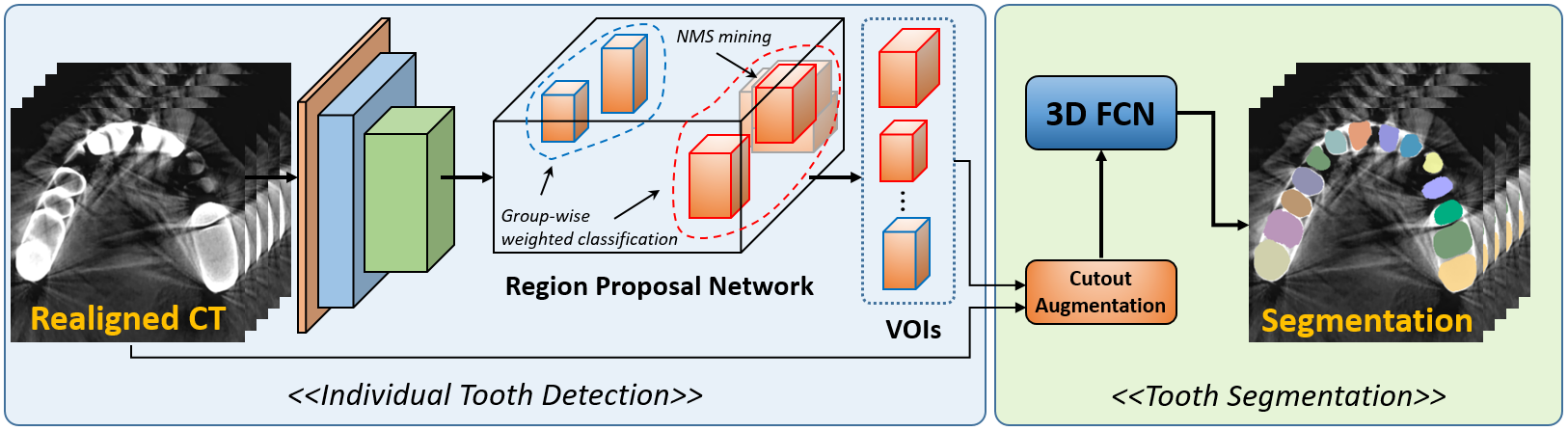}
    \label{fig:archi_ds}}

    \caption{Overall architecture of the proposed method: (a) Workflow of pose-aware image realignment, (b) Subsequent instance segmentation, which performs individual tooth detection and segmentation (Best viewed in color).}
    \label{fig:architecture}
\end{figure*}

\subsection{Object Detection and Instance Segmentation}
There are many elaborate studies on object detection based on CNNs \cite{girshick2015fast, ren2015faster, redmon2016you, chen2018masklab}. The two major approaches are classification- and regression-based approaches. The classification-based object detection method attempts to identify boxes that contain an object by locally windowed classification \cite{ren2015faster, chen2018masklab}. In contrast, the regression-based approach attempts to detect boxes through direct regression \cite{redmon2016you, liu2016ssd, redmon2018yolov3}. The regression-based approaches, such as YoLo algorithm \cite{redmon2016you}, were designed for real-time applications (e.g., automobile, surveillance vision). The classification-based approaches, such as faster R-CNN \cite{ren2015faster}, are superior to the regression-based detection in terms of accuracy primarily because of the fused, local classification and regression procedures. Object detection techniques were successfully employed in the instance segmentation frameworks \cite{he2017mask, chen2018masklab, dai2016instance}. The representative method is Mask-RCNN \cite{he2017mask}, which adapted the faster R-CNN \cite{ren2015faster} framework by extending the network with pixel-wise classification loss at the end (i.e., instance segmentation framework).

\subsection{CNNs in Medical Image Segmentation}
A fully CNN is the most successful method in medical image segmentation tasks \cite{ronneberger2015u, cciccek20163d, chen2017voxresnet, milletari2016v, kamnitsas2017efficient, havaei2017brain, dou20173d, jegou2017one, chen2018deeplab, oktay2018anatomically, gibson2018automatic}. Especially, a full 3D-CNN-based approaches \cite{milletari2016v, chen2017voxresnet, dou20173d, oktay2018anatomically} outperform other 2D-based methods \cite{ronneberger2015u} that are based on 3D convolutions. In \cite{milletari2016v, cciccek20163d}, a full 3D-CNN-based U-net-like architecture was reported to segment volumetric medical images. A dice coefficient loss metric was proposed to overcome the class imbalance issue in \cite{milletari2016v}. A voxelwise residual network, called VoxResNet, performed brain tissue segmentation \cite{chen2017voxresnet}. A residual learning mechanism \cite{he2016deep} was used to classify each voxel in the VoxResNet \cite{chen2017voxresnet}. More recently, training through a shape-prior method was proposed in a densely connected V-net-like structure \cite{gibson2018automatic}.

\section{Methodology}
The proposed individual tooth segmentation method comprises of three steps: VOI realignment, tooth detection, and individual segmentation. The VOI realignment is primarily performed by a pose regression in a projected 2D space (i.e., x-axis-directional projection). The realignment of VOI not only crops the interested region but also aligns the primary axes of the teeth to be better aligned to the image axes. Subsequently, tooth detection and segmentation are performed using CNNs. The details of the algorithm are described in the following subsections.



\subsection{Pose-Aware Volume-of-Interest Realignment}
In the first step, we reduced the dimension of the original 3D image to 2D for a robust detection of the VOI region, as proposed in \cite{chung2019automatic}. Let us consider a computed tomography (CT) image $I(\textbf{x})\longrightarrow\mathbb{R}$ where $\textbf{x}\in\Omega$ ($\Omega\in\mathbb{R}^3$). A maximum intensity projection image corresponding to the x-axis direction, $I_{p}$, is generated from $I$ (Fig. \ref{fig:archi_align}). The image is then normalized to the range of [0-1]. Finally, we used the trained CNN model to acquire the corresponding point and line pairs. Figure \ref{fig:archi_align} shows an example of a single point and line pair for the upper jaw. After the regression of a point and line pair, we cropped the original CT image by a fixed depth value of 12mm corresponding to the line regression (Fig. \ref{fig:archi_align}) to include the entire root area. We applied an additional 2mm of margin in the opposite direction to compensate for potential errors. In the case of the lower jaw, we flipped the realigned image to synchronize the directions of all the teeth to ease the complexity of shape variations.\par

For training, we used $100$ CT images. We manually annotated (i.e., a point and the angle pairs of the lines) images for $I_p$. The overall loss is formulated as follows.
\begin{equation}
\begin{split}
    \chi(X, \mathbf{p}_i, \theta_i; W)=&\sum_{i=0}^{2}{||\mathbf{p}_i-\mathbf{y}_i||}_2+\alpha\sum_{i=0}^{2}{||\theta_i-\phi_i||}_2\\
    &+\beta{||W||}_2^2,
\end{split}
\label{eq:loss1}
\end{equation}

\noindent
where $X$, $\textbf{p}_i$, and $\theta_i$ are the input 2D image, ground-truth 2D points, and angles of the lines, respectively. $W$ represented the weights of the network, $\mathbf{y}_i$ and $\phi_i$, as the network outputs. The network is trained according to the weighting parameters $\alpha$ and $\beta$. For training and inference, we used the traditional VGG-16 model developed by the Visual Geometry Group \cite{simonyan2014very} with a minor modification in the final layer to output a 6D tensor (i.e., 2D point and angle pairs). ``Xavier" initialization \cite{glorot2010understanding} was used for initializing all the weights of the network. While training the network, we fixed the loss parameters as $\alpha=\beta=0.1$. We used the Adam optimizer \cite{kingma2014adam} with a batch size of 32 and set the learning rate to 0.001. We decayed the learning rate by multiplying 0.1 for every 20 epochs. We trained the network for 100 epochs using an Intel i7-7700K desktop system with 4.2 GHz processor, 32 GB of memory, and Nvidia Titan XP GPU machine. It took 1h to complete all the training procedures.

\subsection{Tooth Detection}
Once VOI was realigned, we performed individual tooth detection. We considered a cropped and realigned image, $I_{\mathbf{p}, \theta, d}$, where $\mathbf{p}, \theta$ are a point and an angle regressed by a pose regression step, respectively, and $d$ indicates the depth of the VOI region with respect to the pose aligning vector (Fig. \ref{fig:archi_align}). We employed a depth of 14mm to include all the teeth, based on to the typical size of the anatomical human tooth presented in \cite{nelson2014wheeler}. We adopted the faster R-CNN framework \cite{ren2015faster} as a baseline and applied a few modifications: 1) sampling of the anchors for training the classifier in the region proposal network (RPN) by mining true samples through non-maximum suppression (NMS) and 2) transforming the single-class classification task to a multiclass by anatomical grouping. The former anchor classification in the RPN module is critical for the initial box proposals. To resolve the hard example mining problem while training the classifier, we applied the NMS technique while training the RPN module. That is, NMS was applied both in the training procedure of the RPN module and in the localizing step for the final output. In the latter grouped classification, we transformed a given single-class classification problem into a multiclass form by grouping the teeth based on the anatomical shapes. We used three classes, i.e., metal, canines/premolars (i.e., one rooted; the identified numbers were 11-13, 21-23, 31-33, 41-43), and the others (i.e., two or more rooted). The group-wise weighted classification aided the RPN proposals of metal-teeth, and thus, improved the final accuracy.\par


\begin{figure*}[t]
    \centering
    \includegraphics[width=\linewidth]{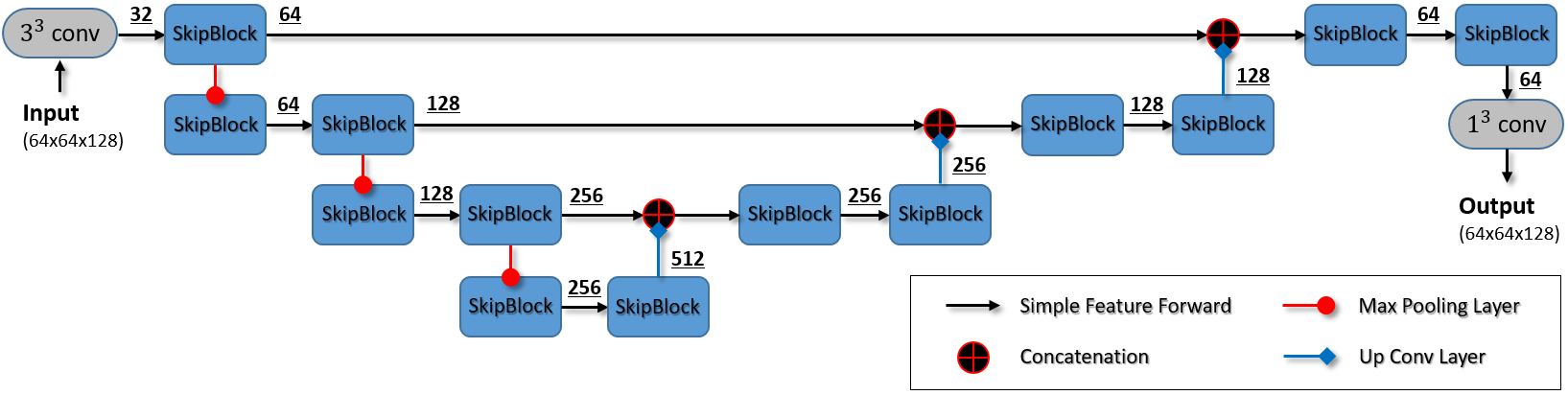}
    \caption{Proposed 3D volumetric FCN architecture. The network has the same architecture as 3D U-net \cite{cciccek20163d}; however, each nonlinear layer (i.e., convolution layer, batch normalization, and rectified linear unit activation function) is replaced by a SkipBlock (Fig. \ref{fig:skipblock}). The blue (squared) and red (circled) arrows indicate the up-convolution and max-pooling layer, respectively, as in the original paper \cite{cciccek20163d}. The numbers demonstrate the output features for each SkipBlock unit.}
    \label{fig:network}
\end{figure*}

Finally, we applied a 2mm margin, i.e., dilation, to each axis of the output boxes to compensate for a possible inaccurate detection. We resized the realigned image to $224\times 224\times 112$ for all inputs of the network. The performance of the tooth detector can be improved significantly by image realignment and VOI cropping to reduce the overlapping ratio of an object. It is clear that reducing the inter-overlapping area boosts the NMS performance for true example mining (i.e., sampling), which leads to accurate region detection. Moreover, the tooth-to-volume ratio significantly increased through VOI realignment; thus, our proposed framework runs without a patch-wise input cropping procedure \cite{cui2019toothnet}, which was previously proposed to resolve the problems that arise from anchor-based methods for small objects. The comparative experiments and ablation studies are described in Section \Romannum{4}.


    
    

\subsection{Individual Tooth Segmentation}
The individual tooth segmentation was performed by a single CNN. We adopted the base architecture of the 3D U-net \cite{cciccek20163d} which is a popular network for medical image segmentation. The proposed network has three significant differences from the 3D U-net \cite{cciccek20163d}, i.e.,: 1) the replacement of a single convolution layer by a skip-connected block that is parameter-efficient, 2) the employment of cutout augmentation \cite{devries2017improved, zhong2017random}, and 3) the modification of the final loss function (i.e., voxel-wise classification) to a distance map regression. The former modification was for the efficiency of the network and the latter two modifications were designed to overcome the presence of severe metal artifacts and inter-tooth proximity on CBCT images, respectively.

\begin{figure}[t]
    \centering
    \includegraphics[width=\linewidth]{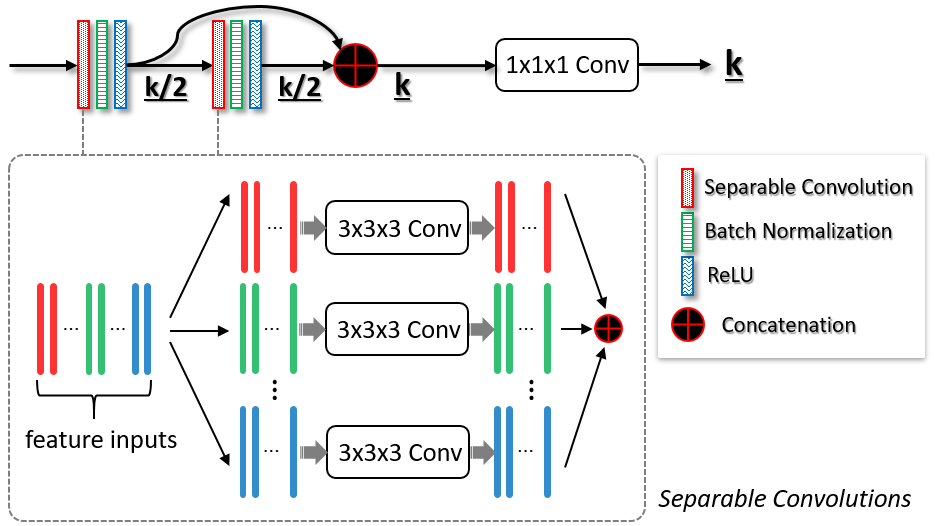}
    \caption{Skip-connected separable convolution block (i.e., SkipBlock). The input features for each separable convolution are divided into groups. All groups are concatenated after separated convolutions. The two nonlinear layers (i.e., separable convolution, batch normalization, and ReLU activation function) were used in a single block. The output features are produced by a $1^3$ convolution applied to concatenated features. $k$ indicates the number of final output features.}
    \label{fig:skipblock}
\end{figure}

\subsubsection{Base Architecture}
Our proposed network (Fig. \ref{fig:network}) architecture has an analysis path and a synthesis path, as in the standard U-net \cite{cciccek20163d}, each with four resolution steps. The down-sampling of the dimension was performed by a $2^3$ max pooling layer with strides of two, and the up-sampling by a $2^3$ transposed convolution (i.e., de-convolution) with strides of two. The network was implemented similar to the 3D U-net \cite{cciccek20163d}, except for the use of SkipBlocks (Fig. \ref{fig:skipblock}). We replaced each single set of the convolution layer, batch normalization, and rectified linear unit with a single SkipBlock. The SkipBlock contained two series of non-linearities and a skip connection. The SkipBlock applied a separable convolution technique \cite{chollet2017xception}, which is known to reduce the number of parameters and improve the performance of generalization. We used four separable groups in all the experiments.\par

\subsubsection{Metal-Intensive Augmentation}
We applied the cutout augmentation \cite{devries2017improved} to improve segmentation results in the metal artifact region. The inference within the metal artifact region, similarly to the inplainting manner, enhanced the final segmentation. The position of the cutout mask was not constrained to the boundaries. A randomly sized zero mask was applied in the range of \(L/5\leq l \leq L/4\), where \(l\) and \(L\) are the lengths of the mask and image in each dimension, respectively.

\subsubsection{Distance Map Regression Loss}
We employed a distance map to train the network. The ground-truth annotated label was first transformed to a distance map. The distance map is defined by assigning the distance to the closest point in the background to each pixel, as presented below:

\begin{equation}
    \bold{l}_D(\mathbf{i})=min_{\mathbf{b}\in B}Dist(\mathbf{i}, \mathbf{b}),
\label{eq:dt}
\end{equation}

\noindent
where $B\subset\mathbf{l}$ is a set of background in the ground-truth label $\mathbf{l}$, $Dist$ is a distance function, and $\mathbf{i}\in\mathbb{R}^3$. We used Chamfer distance \cite{borgefors1986distance} to approximate the Euclidean distance transformation. The overall loss function is defined by mean squared error (MSE) between the distance map and the final output of the network:

\begin{equation}
    \textit{L}(\hat{I_{\mathbf{p}, \theta, d}}, \bold{l}_D;W)=\textit{MSE}(\bold{y}, \bold{l}_D)+\alpha\|W\|_2^2,
\label{eq:loss2}
\end{equation}

\noindent
where $\hat{I_{\mathbf{p}, \theta, d}}$, $\bold{l}_D$, $W$ are the cropped tooth image, distanced transformed ground-truth map (\ref{eq:dt}), and the weights of the network, respectively. $\bold{y}$ is the final output of the network and $\alpha$ is a weighting parameter that controls the impact of $L_2$ regularization.\par

\subsection{Learning the Network}
In total, the manual annotation of 50 subjects (CT images) was acquired with the help of clinical experts in the field. In the dataset, the slice thickness values ranged from 0.2-0.4mm, and the pixel sizes ranged from 0.2-0.4mm. For the training dataset, the cropped individual tooth images were resampled into $64\times64\times128$. The input images were normalized in the range [0-1] for each voxel. On-the-fly random affine deformations were subsequently applied to the dataset for each iteration with 80\% probability. Finally, the proposed cutout image augmentation was performed with an 80\% probability.\par

``Xavier" initialization \cite{glorot2010understanding} was used for initializing all the weights of the proposed network. While training the network, the parameter $\alpha$ was fixed to 0.1 in (\ref{eq:loss2}). The Adam optimizer was used with a batch size of four and a learning rate of 0.001. The learning rate was decayed by multiplying 0.1 for every 50 epochs. The network was trained for 150 epochs using an Intel i7-7700K desktop system with a 4.2 GHz processor, 32 GB memory, and Nvidia Titan XP GPU machine. It took 10 h to complete all the training procedures.

\section{Experiments}
In this section, we present the evaluations of the proposed pose-aware (PA) tooth detector and the subsequent individual tooth segmentation. The experimental dataset included metal-intensive cases, which are commonly observed in dental clinics. The CT images were sourced from a multicenter including four different centers. We used 100 independent subjects for training the pose regression network (\ref{eq:loss1}), and 50 subjects with each tooth defined for training the individual segmentation network (\ref{eq:loss2}). For testing, we used the other 25 subjects for all the quantitative evaluations.

    
    

\begin{figure}[t]
    \centering
    \includegraphics[width=\linewidth]{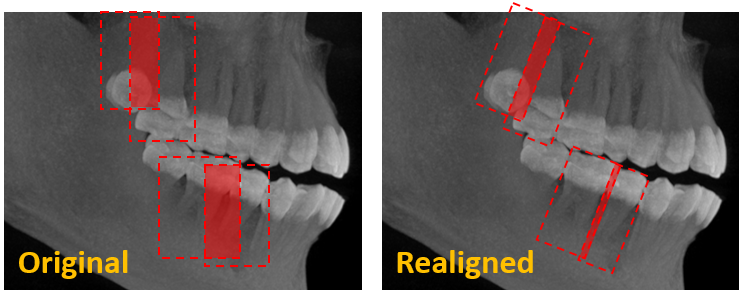}
    \caption{Sample ground-truth box annotations. The realigned boxes (right) show lower overlaps between boxes.}
    \label{fig:vis_box}
\end{figure}

\subsection{Performance of Pose-Aware Detector}
The overlapping ratio (OR) of the ground-truth boxes was significantly reduced owing to the VOI realignment. The overlapping ratio can be calculated by

\begin{equation}
    OR=\frac{A\cap \bar{A}}{A},
\label{eq:overlap}
\end{equation}

\noindent
where $A$ indicates the area of the ground-truth box and $\bar{A}$ indicates the overlapping neighbor boxes. The \textit{mean overlapping ratio} of the ground-truth boxes in the original image alignment was 0.32 while that in the PA VOI realigned image was 0.19. The realignment showed a 38.66\% reduction in the overlapping area between boxes. To evaluate the accuracy of the detection, we newly defined a ratio called object include ratio (OIR), which is a ratio (\%) of the foreground region of an included object in a detected box to the ground-truth foreground:

\begin{equation}
    OIR=\frac{A_o\cap\{A_o\subset\Tilde{A}\}}{A_o},
\label{eq:oir}
\end{equation}

\noindent
where $A_o$ indicates the area of an object inside the ground-truth box ($A$) and $\Tilde{A}$ is the detected box. OIR is a critical measurement of whether the individual tooth region is included inside the box that is detected by a neural network. Table \ref{table:iou} shows the average precision (AP) \cite{lin2014microsoft} with threshold value of 0.5 for an intersection over union (i.e., $AP_{50}$) and the OIR results of the proposed tooth detection (TRCNN). PATRCNN (i.e., pose-aware TRCNN), which employed the realignment, showed significant improvement in the $AP_{50}$. By applying a marginal expansion of the boundaries, the OIR reached up to $0.9951\pm0.0023$, which indicates that the majority of the individual tooth region is included in the detected boxes.\par

\begin{table}[tb]
\renewcommand{\arraystretch}{1.8}
\captionsetup{justification=centering, labelsep=newline}
\caption{AVERAGE PRECISION AND MEAN OBJECT-INCLUDE-RATIO OF THE TOOTH DETECTION}
\label{table:iou}
\begin{tabularx}{\linewidth}{l|>{\centering\arraybackslash}X|>{\centering\arraybackslash}X}
\centering
\textbf{Methods} & $\mathbf{AP_{50}}$ & \textbf{OIR} \\
\hline
TRCNN & $67.88$ & $0.83\pm0.03$\\
PATRCNN & \textbf{89.75} & $0.90\pm0.02$\\
PATRCNN with margin & N/A & \textbf{1.00} $\pmb{\pm}$ \textbf{0.00}\\
\end{tabularx}
\end{table}

\begin{table*}[t]
\renewcommand{\arraystretch}{1.7}
\captionsetup{justification=centering, labelsep=newline}
\caption{QUANTITATIVE RESULTS OF THE STATE-OF-THE-ART TOOTH SEGMENTATION METHODS}
\label{table:results}
\begin{tabularx}{\textwidth}{l||>{\centering\arraybackslash}X|>{\centering\arraybackslash}X|>{\centering\arraybackslash}X|>{\centering\arraybackslash}X|>{\centering\arraybackslash}X|>{\centering\arraybackslash}X}
\textbf{Methods} & \textbf{F1 Score} & \textbf{AJI} & \textbf{Precision} & \textbf{Sensitivity} & \textbf{HD [mm]} & \textbf{ASSD [mm]}\\
\hline
Seed + Levelset \cite{gan2017tooth} & $0.85\pm0.13$ & N/A & $0.86\pm0.07$ & $0.87\pm0.16$ & $3.19\pm2.28$ & $0.46\pm0.86$\\
Seed + Levelset \cite{wang2018accurate} & $0.84\pm0.13$ & N/A & $0.85\pm0.08$ & $0.87\pm0.16$ & $3.51\pm2.17$ & $0.49\pm0.85$\\
\hline
ToothNet \cite{cui2019toothnet} & $0.88\pm0.06$ & $0.66\pm0.07$ & $0.88\pm0.08$ & $0.89\pm0.11$ & $2.75\pm1.49$ & $0.32\pm0.14$\\
\hline
Mask-RCNN \cite{he2017mask} & $0.83\pm0.21$ & $0.47\pm0.18$ & $0.91\pm0.06$ & $0.82\pm0.26$ & $3.47\pm3.22$ & $0.56\pm0.86$\\
\hline
TRCNN+TSNet (ours w/o PA) & $0.90\pm0.12$ & $0.84\pm0.00$ & $0.91\pm0.11$ & $0.91\pm0.14$ & $2.04\pm1.84$ & $0.29\pm0.51$\\
\textbf{PATRCNN+TSNet (ours)} & $\textbf{0.93}\pmb{\pm}\textbf{0.03}$ & $\textbf{0.86}\pmb{\pm}\textbf{0.01}$ & $\textbf{0.93}\pmb{\pm}\textbf{0.04}$ & $\textbf{0.93}\pmb{\pm}\textbf{0.07}$ & $\textbf{1.59}\pmb{\pm}\textbf{1.22}$ & $\textbf{0.20}\pmb{\pm}\textbf{0.10}$\\
\end{tabularx}
\end{table*}

\begin{figure*}[tb]
    \centering
        \begin{minipage}[b]{1.33in}
            \captionsetup[subfigure]{labelformat=parens,labelsep=space,font=small}
            \centering
                \vfil
                \includegraphics[width=1.3in]{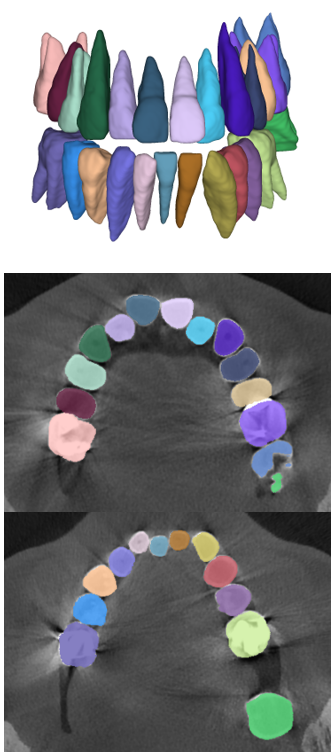}
            \captionof{subfigure}{Ground-truth.}
        \end{minipage}
        \hfil
        \begin{minipage}[b]{1.33in}
            \captionsetup[subfigure]{labelformat=parens,labelsep=space,font=small}
            \centering
                \vfil
                \includegraphics[width=1.3in]{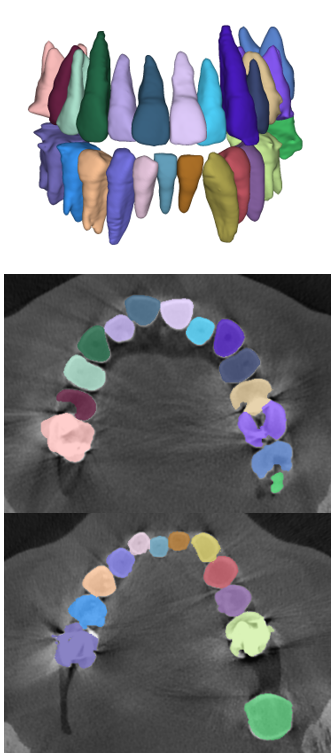}
            \captionof{subfigure}{Level-set \cite{gan2017tooth}.}
        \end{minipage}
        \hfil
        \begin{minipage}[b]{1.33in}
            \captionsetup[subfigure]{labelformat=parens,labelsep=space,font=small}
            \centering
                \vfil
                \includegraphics[width=1.3in]{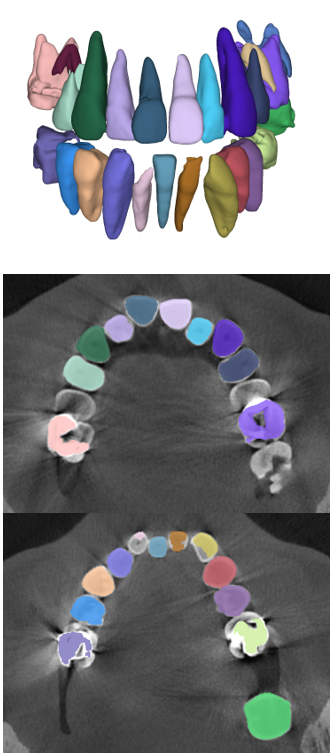}
            \captionof{subfigure}{ToothNet \cite{cui2019toothnet}.}
        \end{minipage}
        \hfil
        \begin{minipage}[b]{1.33in}
            \captionsetup[subfigure]{labelformat=parens,labelsep=space,font=small}
            \centering
                \vfil
                \includegraphics[width=1.3in]{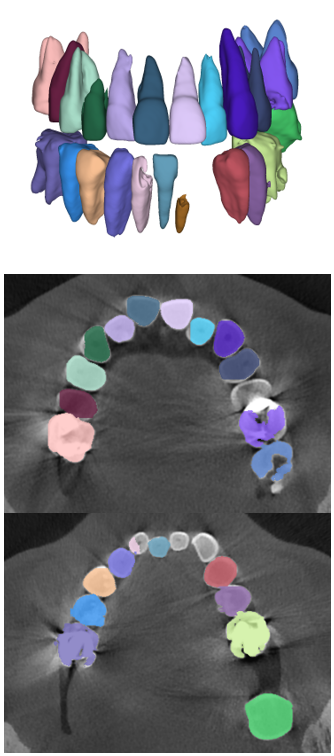}
            \captionof{subfigure}{Mask-RCNN \cite{he2017mask}.}
        \end{minipage}
        \hfil
        \begin{minipage}[b]{1.33in}
            \captionsetup[subfigure]{labelformat=parens,labelsep=space,font=small}
            \centering
                \vfil
                \includegraphics[width=1.3in]{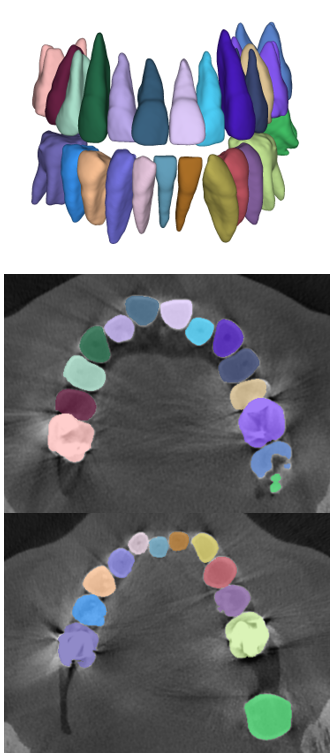}
            \captionof{subfigure}{PATRCNN+TSNet.}
        \end{minipage}
    \caption{Visualization of the segmentation results on the test image: (a) First column shows the ground-truth label annotated by the experts in the field. The columns from the second indicate the following: (b) level-set method \cite{gan2017tooth}, (c) ToothNet \cite{cui2019toothnet}, (d) Mask-RCNN \cite{he2017mask}, and (e) our method, respectively. Note that the level-set method (i.e., the second column) is performed by manually specifying initial contours for each tooth.}
    \label{fig:vis}
\end{figure*}

\begin{figure*}[ht!]
    \centering
    \subfloat[Level-set \cite{gan2017tooth}.]{
    \includegraphics[width=1.6in]{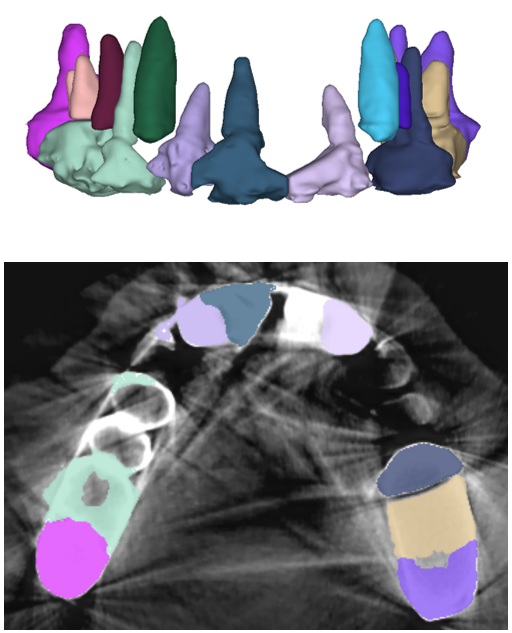}
    \label{fig:vis2_levelset}}
    \hfil
    \subfloat[ToothNet \cite{cui2019toothnet}.]{
    \includegraphics[width=1.6in]{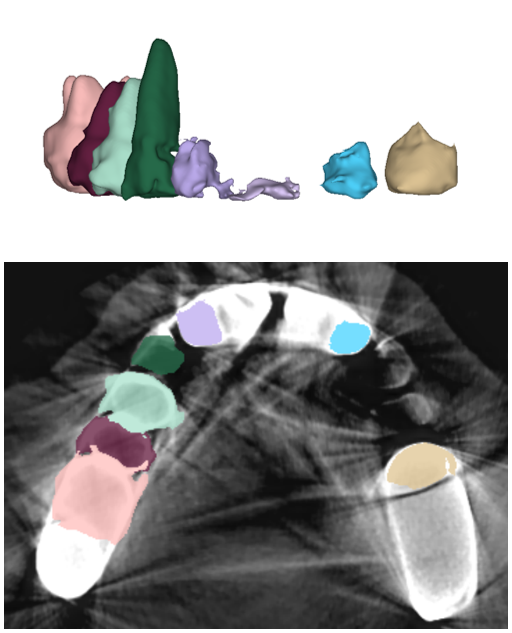}
    \label{fig:vis2_toothnet}}
    \hfil
    \subfloat[Mask-RCNN \cite{he2017mask}.]{
    \includegraphics[width=1.6in]{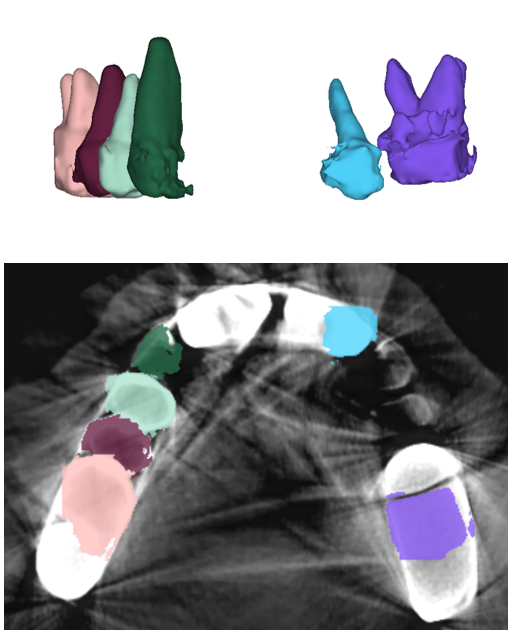}
    \label{fig:vis2_maskrcnn}}
    \hfil
    \subfloat[Ours.]{
    \includegraphics[width=1.6in]{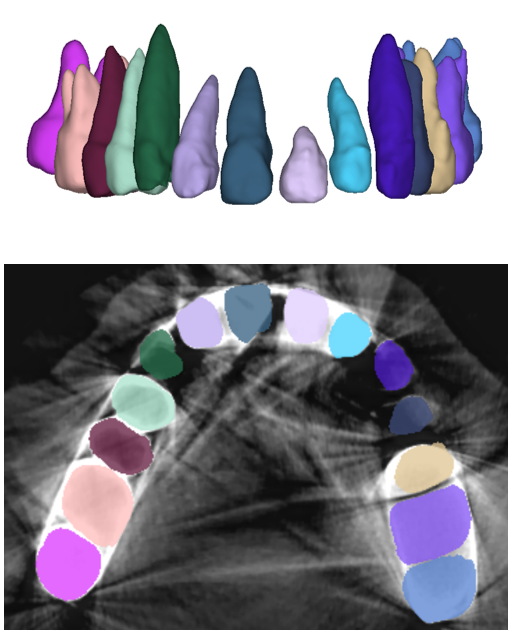}
    \label{fig:vis2_ours}}
    
    \caption{Visualization of the segmentation results on the test image with severe metal artifacts. The example case shows one of the relatively more severe metal artifacts observed in the clinics. Note that the ground-truth label is not obtained, even by the experts, owing to the difficulty of tooth delineation on metal artifacts. (a) The level-set method \cite{gan2017tooth} suffers from accurate propagation of tooth boundaries. The results of (b) ToothNet \cite{cui2019toothnet} and (c) Mask-RCNN \cite{he2017mask} demonstrate that many teeth with metal artifacts are difficult to detect. (d) Our proposed PATRCNN+TSNet shows the best detection and segmentation results.}
    \label{fig:vis2}
\end{figure*}

\begin{table*}[t]
\renewcommand{\arraystretch}{1.7}
\captionsetup{justification=centering, labelsep=newline}
\caption{ASSESSMENT OF SEGMENTATION NETWORKS BASED ON THE GROUND-TRUTH BOXES}
\label{table:results_gt}
\begin{tabularx}{\textwidth}{l||>{\centering\arraybackslash}X|>{\centering\arraybackslash}X|>{\centering\arraybackslash}X|>{\centering\arraybackslash}X|>{\centering\arraybackslash}X}
\textbf{Methods} & \textbf{F1 Score} & \textbf{Precision} & \textbf{Sensitivity} & \textbf{HD [mm]} & \textbf{ASSD [mm]}\\
\hline
\textbf{GT}+3D U-net \cite{cciccek20163d} & $0.89\pm0.03$ & $0.92\pm0.05$ & $0.87\pm0.07$ & $2.71\pm1.26$ & $0.30\pm0.11$\\
\textbf{GT}+VoxResNet \cite{chen2017voxresnet} & $0.93\pm0.02$ & $0.91\pm0.05$ & $0.96\pm0.03$ & $1.06\pm0.60$ & $0.18\pm0.04$\\
\textbf{GT}+DenseVNet \cite{gibson2018automatic} & $0.92\pm0.03$ & $0.97\pm0.03$ & $0.88\pm0.07$ & $1.70\pm1.23$ & $0.21\pm0.08$\\
\textbf{GT}+TSNet (ours w/o PA) & $\textbf{0.93}\pmb{\pm}\textbf{0.03}$ & $\textbf{0.97}\pmb{\pm}\textbf{0.02}$ & $\textbf{0.89}\pmb{\pm}\textbf{0.06}$ & $\textbf{1.08}\pmb{\pm}\textbf{0.67}$ & $\textbf{0.18}\pmb{\pm}\textbf{0.07}$\\
\hline
\textbf{PAGT}+3D U-net \cite{cciccek20163d} & $0.92\pm0.02$ & $0.92\pm0.05$ & $0.91\pm0.05$ & $2.10\pm1.13$ & $0.23\pm0.08$\\
\textbf{PAGT}+VoxResNet \cite{chen2017voxresnet} & $0.94\pm0.02$ & $0.91\pm0.05$ & $0.97\pm0.03$ & $0.93\pm0.54$ & $0.16\pm0.04$\\
\textbf{PAGT}+DenseVNet \cite{gibson2018automatic} & $0.93\pm0.03$ & $0.97\pm0.02$ & $0.89\pm0.06$ & $1.25\pm0.81$ & $0.19\pm0.07$\\
\textbf{PAGT+TSNet (ours)} & $\textbf{0.96}\pmb{\pm}\textbf{0.01}$ & $\textbf{0.96}\pmb{\pm}\textbf{0.03}$ & $\textbf{0.96}\pmb{\pm}\textbf{0.02}$ & $\textbf{0.86}\pmb{\pm}\textbf{0.44}$ & $\textbf{0.15}\pmb{\pm}\textbf{0.04}$\\
\end{tabularx}
\end{table*}

\subsection{Segmentation Performance}
\subsubsection{Evaluation Metric}
The segmentation results were evaluated using the F1 score, aggregated Jaccard index (AJI), precision, sensitivity, Hausdorff distance (HD), and average symmetric surface distance (ASSD). The F1 score is defined as follows:

\begin{equation}
    F_1=2\times\frac{Precision\times Sensitivity}{Precision+Sensitivity}.
\label{eq:f1}
\end{equation}

\noindent
Precision and sensitivity are defined by $P=\frac{TP}{TP+FP}$ and $S=\frac{TP}{TP+FN}$, respectively, where TP, FN, and FP are the numbers of true positive, false negative, and false positive voxels, respectively. The F1 score is equivalent to the dice coefficient \cite{milletari2016v}. The AJI metric is a per-object metric, as presented in \cite{kumar2017dataset}. The one-to-one correspondences were first matched between the ground-truth box and the one detected by maximizing the Jaccard index. Then, the AJI was calculated similar to the Jaccard index; however, the falsely detected components were added to the denominator \cite{naylor2018segmentation}:

\begin{equation}
    AJI=\frac{\sum_{i}^{}|G_i\cap S_k^*(i)|}{\sum_{i}^{}|G_i\cup S_k^*(i)|+\sum_{l\in U}^{}|S_l|},
\label{eq:aji}
\end{equation}

\noindent
where $|\cdot|$ is the cardinality of a set. $G_i$ and $S_k^*(i)$ are the ground-truth individual object voxels and the object voxels by the corresponding box, which are detected by maximizing the Jaccard index (i.e., $S_k^*(i)=\{\argmax_{S_i}\frac{|G_i\cap S_i|}{|G_i\cup S_i|}\}$), respectively, and $U$ is the set of indices of false detected components that were retained by matching the set, $S_k^*(i)$.\par

The surface distance metrics were evaluated on the integrated teeth basis. Let $\textbf{S}_X$ be a set of surface voxels of a set $X$; then, the shortest distance of an arbitrary voxel $p$ can be defined as follows:
\begin{equation}
    d(p, \textbf{S}_X)=\min_{s_X \in \textbf{S}_X}{||p-s_X||}_2.
\label{eq:d}
\end{equation}
Thus, HD is defined as follows \cite{heimann2009comparison}:
\begin{equation}
    HD(X,Y)=\max\{\max_{s_X \in \textbf{S}_X}{d(s_X, \textbf{S}_Y)}+\max_{s_Y \in \textbf{S}_Y}{d(s_Y, \textbf{S}_X)}\}.
\label{eq:hd}
\end{equation}
The distance function is defined as:
\begin{equation}
    D(\textbf{S}_X, \textbf{S}_Y)=\Sigma_{s_X \in \textbf{S}_X} d(s_X, \textbf{S}_Y),
\end{equation}
Moreover, the ASSD can be defined as follows \cite{heimann2009comparison}:
\begin{equation}
    ASSD(X,Y)=\frac{1}{|\textbf{S}_X|+|\textbf{S}_Y|}(D(\textbf{S}_X, \textbf{S}_Y)+D(\textbf{S}_Y, \textbf{S}_X)).
\end{equation}
\par

\subsubsection{Comparison}
The overall evaluation of segmentation performances is presented in Table \ref{table:results}. We first demonstrate manual seeded (i.e., contoured) level-set-based methods \cite{gan2017tooth, wang2018accurate} and ToothNet \cite{cui2019toothnet}, which are state-of-the-art methods for individual tooth segmentation. Our proposed framework, together with the Mask-RCNN \cite{he2017mask}, are presented to show the effectiveness of the PA detection and the subsequent tooth segmentation.\par

\begin{table}[tb]
\renewcommand{\arraystretch}{1.8}
\captionsetup{justification=centering, labelsep=newline}
\caption{PERFORMANCE OF PA DETECTORS}
\label{table:ablation_trcnn}
\begin{tabularx}{\linewidth}{l|>{\centering\arraybackslash}X|>{\centering\arraybackslash}X}
\centering
\textbf{Methods} & $\mathbf{AP_{50}}$ & \textbf{OIR} \\
\hline
PA-Faster R-CNN \cite{ren2015faster} & $72.20$ & $0.86\pm0.03$\\
\hline
PATRCNN & \textbf{89.75} & \textbf{0.90} $\pmb{\pm}$ \textbf{0.02}\\
PATRCNN w/o NMS sampling & $76.72$ & $0.89\pm0.02$\\
PATRCNN w/o grouping & $84.59$ & $0.88\pm0.03$\\
\end{tabularx}
\end{table}

\begin{table}[tb]
\renewcommand{\arraystretch}{1.8}
\captionsetup{justification=centering, labelsep=newline}
\caption{PERFORMANCE OF TSNet AND ITS VARIANTS}
\label{table:ablation_tsnet}
\begin{tabularx}{\linewidth}{l|>{\centering\arraybackslash}X|>{\centering\arraybackslash}X|>{\centering\arraybackslash}X}
\centering
\textbf{Methods} & \textbf{F1 Score} & \textbf{HD [mm]} & \textbf{ASSD [mm]} \\
\hline
PAGT+TSNet & $\textbf{0.96}\pmb{\pm}\textbf{0.01}$ & $\textbf{0.86}\pmb{\pm}\textbf{0.44}$ & $\textbf{0.15}\pmb{\pm}\textbf{0.04}$\\
PAGT+TSNet-C & $0.94\pm0.02$ & $0.90\pm0.48$ & $0.16\pm0.03$\\
PAGT+TSNet-D & $0.91\pm0.04$ & $0.96\pm0.51$ & $0.24\pm0.07$\\
PAGT+TSNet-CD & $0.92\pm0.04$ & $0.96\pm0.50$ & $0.22\pm0.07$\\
\end{tabularx}
\end{table}

\begin{figure}[t]
    \centering
    
    \subfloat[F1 score.]{
    \includegraphics[width=\linewidth]{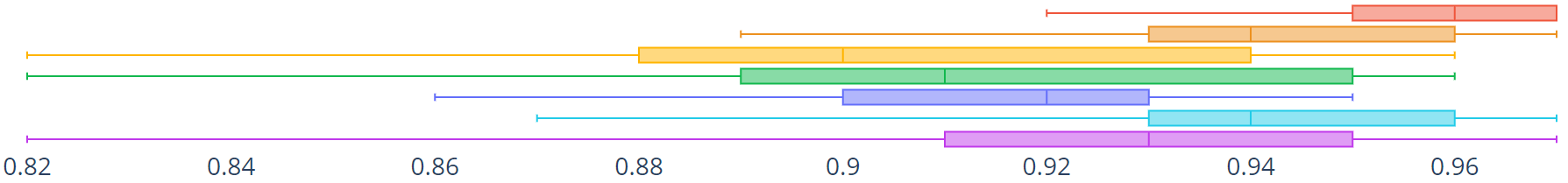}
    \label{fig:plot_f1}}
    \vfil
    \subfloat[Precision.]{
    \includegraphics[width=\linewidth]{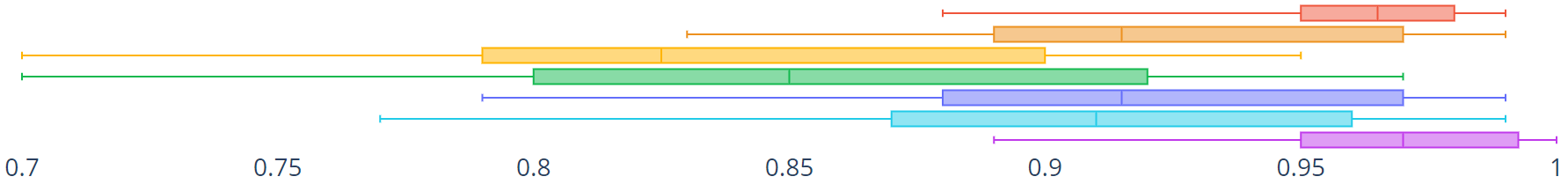}
    \label{fig:plot_precision}}
    \vfil
    \subfloat[Sensitivity.]{
    \includegraphics[width=\linewidth]{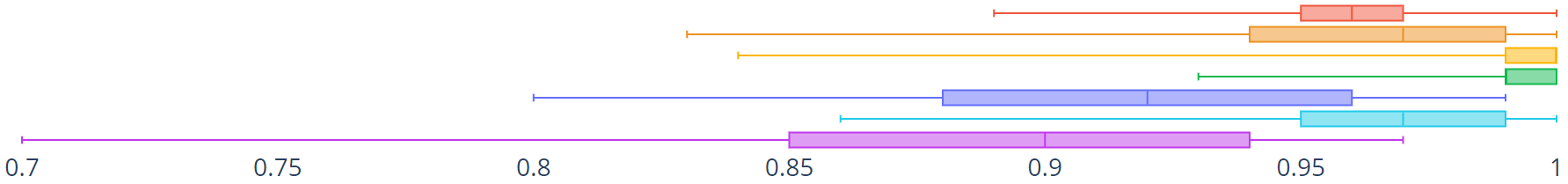}
    \label{fig:plot_sensitivity}}
    \vfil
    \subfloat[HD.]{
    \includegraphics[width=\linewidth]{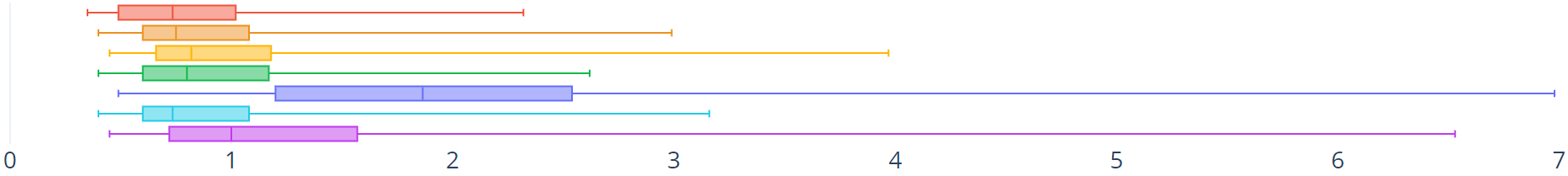}
    \label{fig:plot_hd}}
    \vfil
    \subfloat[ASSD.]{
    \includegraphics[width=\linewidth]{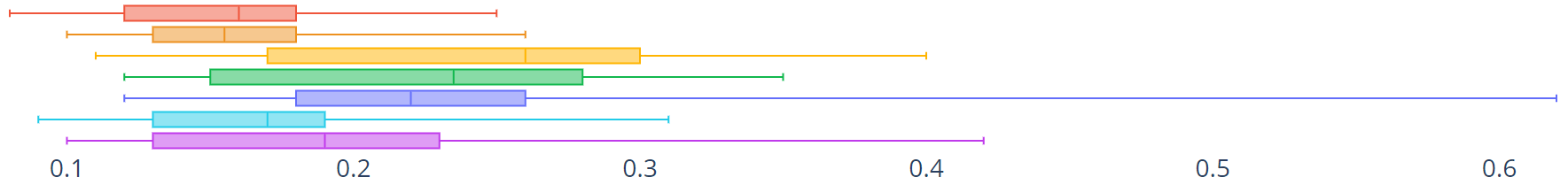}
    \label{fig:plot_assd}}
    \vfil

    \includegraphics[width=\linewidth]{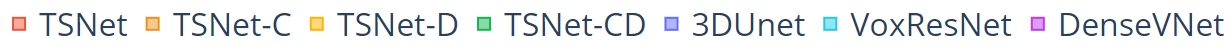}
    \caption{Box plots of PAGT-based segmentation methods.}
    \label{fig:plot}
\end{figure}

\begin{figure*}[t]
    \centering
    \subfloat[Ground-truth.]{
    \includegraphics[width=0.7in]{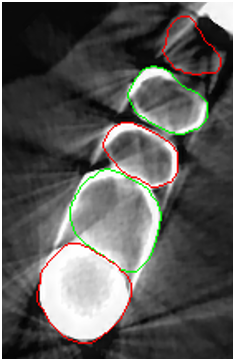}
    \label{fig:vis_metal_gt}}
    \hfil
    \subfloat[TSNet (ours).]{
    \includegraphics[width=0.7in]{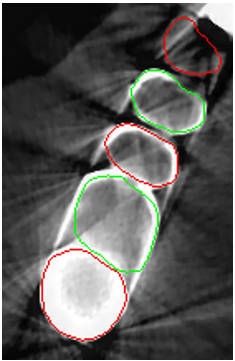}
    \label{fig:vis_metal_tsnet}}
    \hfil
    \subfloat[TSNet-C.]{
    \includegraphics[width=0.7in]{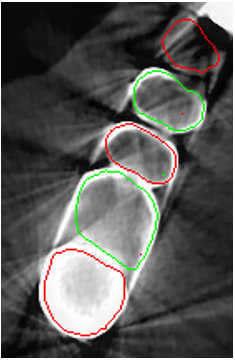}
    \label{fig:vis_metal_tsnet-c}}
    \hfil
    \subfloat[TSNet-D.]{
    \includegraphics[width=0.7in]{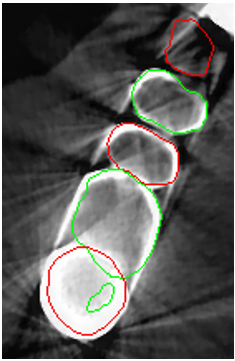}
    \label{fig:vis_metal_tsnet-d}}
    \hfil
    \subfloat[TSNet-CD.]{
    \includegraphics[width=0.7in]{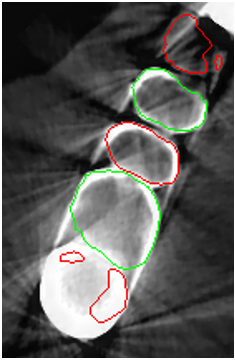}
    \label{fig:vis_metal_tsnet-cd}}
    \hfil
    \subfloat[3D U-net.]{
    \includegraphics[width=0.7in]{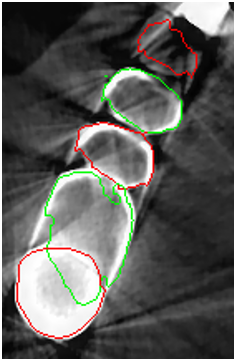}
    \label{fig:vis_metal_3D U-net}}
    \hfil
    \subfloat[VoxResNet.]{
    \includegraphics[width=0.7in]{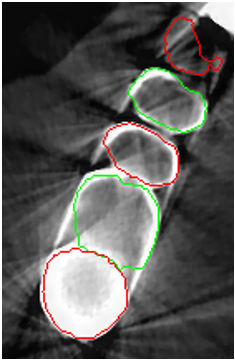}
    \label{fig:vis_metal_voxresnet}}
    \hfil
    \subfloat[DenseVNet.]{
    \includegraphics[width=0.7in]{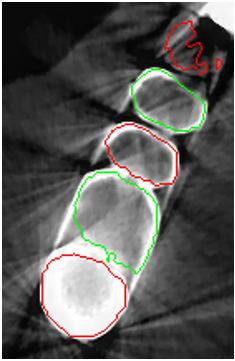}
    \label{fig:vis_metal_densevnet}}
    
    
    \caption{Visualization of the PAGT-based networks.}
    \label{fig:vis_metal}
\end{figure*}

The manual seeded level-set methods \cite{gan2017tooth, wang2018accurate} demonstrate inferior performances while segmenting teeth with predominant metal artifacts (Table \ref{table:results}). The energy-based (i.e., level-set function) algorithmic approach failed while defining an optimum stopping criterion (Fig. \ref{fig:vis}b). The AJI measures for level-set methods are omitted because the box detectors do not exist in the level-set-based methods. In particular, the low precision and sensitivity values indicate that level-set methods have over-segmented or under-segmented the teeth in many cases (Figs. \ref{fig:vis}b and \ref{fig:vis2_levelset}). The accuracies of the Mask-RCNN \cite{he2017mask} and ToothNet \cite{cui2019toothnet} showed comparable performance to the level-set-based methods. The instance segmentation framework successfully detected and segmented the teeth automatically. However, it failed to detect all the teeth regions accurately, which resulted in a degradation of the segmentation performance. Moreover, the performance of the segmentation itself also showed low accuracy owing to the metal artifacts (Figs. \ref{fig:vis}c and \ref{fig:vis}d). Figure \ref{fig:vis2} illustrates the more severe conditions of metal artifacts. The performance of the detection and segmentation processes significantly degraded in ToothNet \cite{cui2019toothnet} (Fig. \ref{fig:vis2_toothnet}) and Mask-RCNN \cite{he2017mask} (Fig. \ref{fig:vis2_maskrcnn}). Conversely, our proposed method outperformed the other state-of-the-art methods (Figs. \ref{fig:vis}e and \ref{fig:vis2_ours}); further, the comparison between the PA- and non-PA-based results also demonstrated that the employment of a PA detector significantly improved the proposed architecture (Table \ref{table:results}). The superior AJI value clearly shows that successful detection improved the overall performance.\par

Table \ref{table:results_gt} shows that our proposed TSNet is superior to other networks in a stand-alone segmentation performance (i.e., using the ground-truth boxes). Moreover, the overall performance of the networks that utilized PA-based ground-truth (PAGT) showed better accuracy than the ones that used the original image axes. The original 3D U-net \cite{cciccek20163d} and others failed to segment the teeth with metal artifacts (Figs. \ref{fig:plot} and \ref{fig:vis_metal}). The DenseVNet \cite{gibson2018automatic} showed relatively higher precision than the 3D U-net \cite{cciccek20163d} and VoxResNet \cite{chen2017voxresnet} as DenseVNet employed a shape-prior based on trainable parameters \cite{gibson2018automatic}. The trained shape-prior based on well-bounded images suppressed the false positive responses. However, DenseVNet demonstrated inaccurate segmentation (i.e., sensitivity) due to the large shape variance of teeth.

    
    

\subsubsection{Ablation Studies}
The performance of our proposed detector (TRCNN) and its variants are listed in Table \ref{table:ablation_trcnn}. The PATRCNN without NMS sampling indicates that while training the anchor classifier, we applied a top-k sampling metric with random negative sampling, which is common for an instance segmentation framework. In the case of PATRCNN without grouping, a single-class classification framework was used for all the classifiers in the network. The result demonstrated that the most significant improvement in the accuracy was obtained through true example mining based on the NMS sampling. The multiclass classification metric by grouping also aided accurate detection. The PA-Faster R-CNN method showed a marginal improvement when compared to TRCNN (in Table \ref{table:iou}), which indicates that the PA method indeed enhances the performance of the detector and neither NMS sampling nor the multiclass classification method can improve the detection without the PA mechanism.\par

We extended the evaluation of segmentation network with the following variants (Table \ref{table:ablation_tsnet}): TSNet-C, TSNet-D, and TSNet-CD, which represent without cutout augmentation, without distance loss (\ref{eq:loss2}), and without both the cutout and the distance loss (i.e., 3D U-net \cite{cciccek20163d} with SkipBlocks), respectively. We used the dice loss in TSNet-D and TSNet-CD. The result shows that the distance loss metric was the primary factor to achieve success in the individual tooth segmentation. Figure \ref{fig:plot} illustrates the box plots of the PAGT-based results and the ablations (Tables \ref{table:results_gt} and \ref{table:ablation_tsnet}) of the proposed network; further, Fig. \ref{fig:vis_metal} illustrates that the proposed TSNet is superior to other methods.

\section{Discussion and Conclusion}

The presence of metal artifacts in CBCT images, which is prevalent condition in clinical practices, hinders the accurate detection and segmentation of teeth. The proposed instance segmentation architecture overcame the challenges by realigning the VOI, improving the detector, and reformulating the segmentation task (i.e., pixel-wise classification) into a distance map regression. The proposed segmentation network focused on identifying metal artifacts and proximate objects. Moreover, the fully automated framework neither required manual inputs nor priors. The instance segmentation framework also demonstrated the advantage of avoiding the difficult pose-based procedure of separating individual teeth.\par

The pose regression-based VOI extraction and realignment method aided the performance of the tooth detector, and also achieved accurate individual tooth segmentation results. The performance of the tooth detector improved by a significant margin owing to two major factors: 1) reduction in the overlapping ratio (\ref{eq:overlap}) between ground-truth boxes and 2) formation of anatomical groups for multiclass classification. A reduced overlapping ratio significantly improved the NMS-based sampling in the true example mining stage, which resulted in detector improvement. By employing NMS-based true example sampling, we acquired accurate region proposals without the difficulty of hard example mining. The multiclass classification framework boosted the accuracy of metal-teeth classification, which also resulted in the improvement of tooth detection. Additionally, the VOI realignment enhanced the tooth-to-volume ratio (TVR) of the input image. The enhancement of TVR demonstrated a huge benefit for instance segmentation because the anchor-based RPN module demonstrates a high probability of ignoring small objects. It is more critical to the volumetric input images owing to the limitation of GPU memories. Unlike \cite{cui2019toothnet}, our proposed network did not require a tedious procedure of cropping the input images while training. The proposed pose regression framework suggests that obtaining a simple cue (i.e., pose) can ease the complexity of the original task in many aspects.\par


\ifCLASSOPTIONcaptionsoff
  \newpage
\fi

\bibliographystyle{IEEEtran}
\bibliography{MyBiB}

\end{document}